\documentclass[conference]{IEEEtran}
\IEEEoverridecommandlockouts
\usepackage{cite}
\usepackage{amsmath,amssymb,amsfonts}
\usepackage{algorithmic}
\usepackage{graphicx}
\usepackage{textcomp}
\usepackage{tabularx, booktabs}

\usepackage{tikz}
\usepackage{pgfplots}
\usepackage{import}
\usepackage{mathbbol}
\usepackage{bbold}
\usepackage{amsmath,amsfonts,amssymb}
\usepackage{float}
\usepackage{environ}
\usepackage{pgfplotstable}
\usepackage{graphicx}
\usepackage{setspace}
\usepackage{hyperref}
\usepackage{soul}

\usepackage{balance}
\usepackage{xcolor}
\usepackage[colorinlistoftodos]{todonotes}
\usepackage{balance}
\def\BibTeX{{\rm B\kern-.05em{\sc i\kern-.025em b}\kern-.08em
		T\kern-.1667em\lower.7ex\hbox{E}\kern-.125emX}}
\begin{document}
	
	\title{Information Plane Analysis Visualization in Deep Learning via Transfer Entropy\\
	}
	
	\author{
		\IEEEauthorblockN{Adrian Moldovan\IEEEauthorrefmark{1}\IEEEauthorrefmark{2},
			Angel Ca\c taron\IEEEauthorrefmark{1}\IEEEauthorrefmark{2},
			R\u azvan Andonie\IEEEauthorrefmark{3}\IEEEauthorrefmark{1}}
		
		\IEEEauthorblockA{\IEEEauthorrefmark{1}\textit{Department of Electronics and Computers}, Transilvania University, Bra\c{s}ov, Romania
		}
		
		\IEEEauthorblockA{\IEEEauthorrefmark{2}
			Siemens Technology, Siemens SRL, Bra\c sov, Romania
		}
		
		\IEEEauthorblockA{\IEEEauthorrefmark{3}Central Washington University, Ellensburg, WA, USA\\
			Email: adrian.moldovan@gmail.com, cataron@unitbv.ro, razvan.andonie@cwu.edu
		}
	}

	\maketitle
	
	\begin{abstract}
		In a feedforward network, Transfer Entropy (TE) can be used to measure the influence that one layer has on another by quantifying the information transfer between them during training. According to the Information Bottleneck principle, a neural model's internal representation should compress the input data as much as possible while still retaining sufficient information about the output. Information Plane analysis is a visualization technique used to understand the trade-off between compression and information preservation in the context of the Information Bottleneck method by plotting the amount of information in the input data against the compressed representation. The claim that there is a causal link between information-theoretic compression and generalization, measured by mutual information, is plausible, but results from different studies are conflicting. In contrast to mutual information, TE can capture temporal  relationships between variables. To explore such links, in our novel approach we use TE to quantify information transfer between neural layers and perform Information Plane analysis. We obtained encouraging experimental results, opening the possibility for further investigations.
		
	\end{abstract}
	
	
	\section{Introduction}
	
	Transfer Entropy (TE) is a statistical measure that is commonly used to quantify the degree of coherence between events, usually those represented as time series. This measure was introduced by Schreiber~\cite{Schreiber2000} and has been linked by some authors \cite{Barnett2009, Hlavackova-Schindler2011} to Granger's causality. Using the term "causality" alone is a misnomer. To avoid further confusion, Granger himself used in 1977 the term "temporally related" \cite{granger1977}. Causality is concerned with whether interventions on a source can have an impact on the target, while information transfer relates to whether observations of the source can aid in predicting state transitions on the target \cite{Lizier2010}. While TE may indicate temporal relationships between two variables, it is not a definitive test for causality, and care should be taken when interpreting the results of TE analysis in this context.
	
	In the case of a multilayered neural network, we can use TE to measure the volume of information that is transferred between the layers as they are successively computed during the network's training process. The temporal relationship between the layers is straightforward: the output of one layer can be seen as the "cause" of the subsequent layer's input, or "effect". This directional information transfer can be measured using TE, which provides a way to analyze the interactions between the different layers of the network. 
	
	The Information Bottleneck (IB) principle \cite{Tishby2001} is another information-theoretical tool that has been used recently to analyze neural models. This principle is founded on the notion that an effective internal representation created by a neural network should maximize compression of the input data and yet preserve correct information about the output.
	
	Shwartz-Ziv and Tishby \cite{ShwartzZiv2017} drew a visualization of the IB using the Information Plane (IP), where IB consists of $I(X ; T)$ and $I(Y ; T)$ as coordinates. $X$ stands for the input, $T$ for the compression, $Y$ represents the output, and $I(\, \cdot \,;\, \cdot \,)$ is the Mutual Information (MI). They analyzed the IP of a deep neural network over time and found that the network undergoes a transition of phases during training, from fitting to compression, where the flow of information becomes more focused and efficient. The two successive distinct phases, \emph{fitting} and \emph{compression} are characterized by $I(X ; T)$, respectively $I(Y ; T)$.
	
	Model discovery, development, verification, and interpretation has been recently enriched with new visual methods having numerous advantages over other alternatives \cite{Kovalerchuk2022}. The human component can be a significant part of this activity since visuals can naturally support explanations of neural models. Up until now, most visualization techniques used to gain insight into deep neural models have focused on visualizing features or generating heatmaps. IP visualization uses a new approach and integrates well into the emerging field know as Visual Knowledge Discovery, combining advances in artificial intelligence, machine learning, and visual analytics \cite{Kovalerchuk2022integrating}.
	
	IP analysis can be used to study deep neural networks to understand the dynamics of information flow and the relationship between input, hidden, and output layers \cite{Cheng2019, Saxe2019}. It has promising potential for deep learning visualization because it can help identify how a model is learning and where it may be struggling. Scatter plots are typically used to visualize an IP. The shape and distribution of the points in the IP can provide insights into the behavior of the model. 
	
	Several authors have performed IP analyses, applied to different NN architectures. For instance, Cheng \emph{et al.} \cite{Cheng2019} used IP analysis to evaluate and select better deep neural networks, with respect to accuracy and training time, for image classification. The outcomes of their study indicate that the IP is a more informative metric than the loss curve, as models exhibiting similar loss curves, which decrease over training epochs, exhibit differing behavior in the IP. 
	
	It has been hypothesized that the evolution of the IP trajectories respects the following dynamics: the generalization performance\footnote{Generalization refers here to the performance of the classifier on the test data and measures the model's ability to adapt properly to new, previously unseen data, drawn from the same distribution as the one used to create the model.} of deep networks is determined in the compression phase. Stochastic gradient descent diffusion-like behavior \cite{Saxe2019}, \cite{Fjellstrom2022} controls the compression phase. This behavior is known to be effective for improved generalization when training neural networks as also shown in Zhu \cite{zhu2018}. Saxe \emph{et al.} \cite{Saxe2019} showed that these claims are not generally true, and reflect instead assumptions made to compute a finite MI metric in deterministic networks. According to \cite{Geiger2022, Saxe2019}, there is no clear causal relationship between compression and generalization in neural networks. Networks that do not compress can still achieve good generalization, and conversely, networks that do compress are not necessarily better at generalization.
	
	Although the IB principle and the idea of a causal link between information-theoretic compression and generalization appear convincing, the results of many experiments conflict with this view. Geiger's recent study \cite{Geiger2022} suggested that compression in the IP does not necessarily indicate the learning of a minimum sufficient statistic, and that this concept may not be essential for good generalization performance. Therefore, the suitability of MI as a measure of a causal relationship is in question, which prompted our current research.
	
	MI and TE capture different aspects of the relationship between two variables. MI measures the amount of information shared between two variables, whereas TE measures the direction and strength of information flow between two variables. TE can capture temporal relationships between variables, whereas MI cannot. Since a temporal  connection between information-theoretic compression and generalization is plausible, using the TE instead of the MI is a promising approach.
	
	The main contribution of our work is to use the TE measure to quantify information transfer between neural layers in IP analyses of neural networks. Our goal is to gain a deeper understanding of the learning dynamics and investigate if there is a connection between compression in the IP and generalization performance. To the extent of our knowledge this is the first time that the TE measure has been applied to demonstrate the IB principle.
	
	The rest of the paper is structured as follows. Section \ref{related} summarizes essential results related to the application of TE and IB in neural networks. Section \ref{IB-TE} introduces our novel model, whereas Section \ref{results} presents experimental results and their interpretations. Section \ref{conclusion} contains the final remarks.
	
	\section{Related Work} \label{related}
	In this section, we provide a summary of important findings related to the use of TE and IB in artificial neural networks.
	
	\subsection{Transfer Entropy in Neural Networks}
	TE is an asymmetric measure of data connectivity and interchange between time series $I$ and $J$, and it is calculated with Schreiber's formula \cite{Schreiber2000}:
	\begin{equation}\label{eq:TEcond}
		TE_{J\rightarrow I}=\sum_{t=1}^{n-1}{p(i_{t+1},i_{t}^{(k)},j_{t}^{(l)}) \: log \frac{p(i_{t+1}|i_{t}^{(k)},j_{t}^{(l)})}{p(i_{t+1}|i_{t}^{(k)})}} 
	\end{equation}
	
	Recently, we can observe a growing interest in applying TE in quantifying the effective connectivity between artificial neurons \cite{Feraud2002, Lizier2011, Vicente2011, Shimono2015, HuiFang2018}. Some results report applications of TE in training neural networks \cite{Patterson2017, Herzog2017, Herzog2020, Obst2010}. 
	
	For instance, in \cite{Herzog2017}, TE is applied to structure the feedback between layers (adjacent or distant) of Convolutional Neural Networks (CNNs) with reduced number of layers. The TE was used to adjust the feedback and was calculated by averaging the values collected from connected neurons. 
	
	Herzog \emph{et al.} \cite{Herzog2020} aimed to provide clear guidelines for computing TE-based neural feedback connectivity in feedforward neural networks, with the goal of improving classification performance. Because of the large number of possible connections, structuring feedback connections in a deep neural model can be very challenging. The authors reduced feedback connections on an AlexNet network down to 3.5\% of all connections, then combined a genetic algorithm to moderately increment the strengths of the retained connections, in similar ways to how the brain amplifies existing feedforward paths. The algorithmic steps involved training the network with standard backpropagation, fine-tuning the network with feedback TE connections, and generating the best performing network using a genetic algorithm. This method improved classification performance.
	
	In previous work \cite{Moldovan2020}, we incorporated the computed TE into the backpropagation algorithm of simple feedforward networks as a scaling factor for gradients. As a result, we reduced the training time and improved the stability of the model. Afterward, we extended the application of TE to include CNNs for the purpose of accelerating the training phase and achieving the desired level of accuracy in fewer epochs \cite{Moldovan2021}. TE served as a smoothing factor that fosters stability by only becoming active periodically, rather than after processing each input sample. 
	
	\subsection{Information Bottleneck in Neural Networks}
	IB is a subset of the information distortion metrics \cite{shannon1959,Dimitrov2001,Equitz1991}. Information distortion is a qualitative metric of the quantization success or of the information loss. Various experiments and theoretical initiatives attempted to investigate the compressed content \emph{T} as a measurable proof of the information that is being encoded during the training processes. Others have simply looked at the distributions of $X$, $Y$ and $T$, either individually, in various pairs or as Markov chains. Pattern recognition problems are predicting the $Y$ outputs by looking at a training set $X$ - the inputs. IB concentrates on the "bottleneck" variable $T$ that aims to maximize the prediction of $Y$ by $I(Y ; T)$ while learning $X$. 
	
	The first significant work that scrutinized IB in the machine learning field was done by Tishby \emph{et al.} \cite{Tishby2000}, for discrete values of $X$, $Y$ and $T$. It was followed by the application of IB objective functions to neural networks \cite{Tishby2015}. In essence, minimizing the following expression aims on maintaining a qualitative compression while improving the accuracy of the decoded variable $Y$: 
	\begin{equation}
		\label{eq:ib}
		min_{P_{T|X}}(I(X;T)-\beta I(Y;T))
	\end{equation}\cite{Tishby2000},  
	where $\beta$ is a Lagrange multiplier that controls the importance of the compressed information captured by $T$.
	
	Goldfeld \emph{et al. } \cite{Goldfeld2019} and Mu\c sat \emph{et al.} \cite{Musat2022} are a few of the recent works that attempted to understand deep learning models using the IB principle. Overviews of IB theory and IP analyses and applications in neural networks were developed by Goldfeld \emph{et al.} \cite{Goldfeld2020} and Geiger \cite{Geiger2022}.
	
	The work of Tishby \emph{et al.} \cite{Tishby2015} was extended by Shwartz-Ziv \emph{et al.} \cite{ShwartzZiv2017} with new findings of the IP on deep networks. They showed that the SGD optimization consists of a first shorter fitting phase and a longer compression phase. 
	
	Saxe \emph{et al.} \cite{Saxe2019} reduced the boundaries stated by Tishby \emph{et al.} in \cite{Tishby2015} to a smaller subset of neural networks, simply by providing eloquent counterexamples. They presented scenarios where generalization is correlated with the activation function and not to IB (i.e., \emph{tanh} activation function provides better compression rates than \emph{ReLU} w.r.t. IB). Another finding of Saxe \emph{et al.} was that SGD produces the same accuracy in the context of IB when compared with regular GD. Probably one of the most important findings from Saxe \emph{et al.} \cite{Saxe2019} is that the compression is observed only in the classification (fully connected) layers. Similar observations were made by Amjad \emph{et al.} \cite{Amjad2020}. 
	
	Goldfeld \emph{et al.} \cite{Goldfeld2018} showed that the reduction (compression) of $I(X ; T)$ is a clustering of training samples based on class. 
	
	Wieczorek \emph{et al.} \cite{Wieczorek2020} created class-level visualization of MI for each layer, revealing the implications of IB for various datasets.
	
	
	Later, Elad \emph{et al.} \cite{Elad2019} used a modified IB loss function to sequentially train each layer. The authors attempted to validate the claim that the maximum IB Lagrangian on each layer offers similar performance with the regular end to end DNN training using cross entropy loss.
	
	Enforcing a lower bound for the IB on a maximization objective function by using variational inference has been successfully experimented by Alemi \emph{et al.} \cite{Alemi2016}, by generating models more robust to overfitting. Kolchinsky \emph{et al.} \cite{Kolchinsky2019} improved Alemi's \emph{et al.} \cite{Alemi2016} lower bound, making it non-parametric and allowing both discrete and continuous $X$ and $Y$. 
	
	Gabri'e \emph{et al.} \cite{Gabrie2019} did not discovered a connection between IB and generalization, when analyzing entropy and MI throughout the training process.
	
	The IB method provides several techniques for determining latent variables that measure the model's comprehension. Saxe \emph{et al.} \cite{Saxe2019} measured the variances of the MI between inputs, labels and compression. They considered binning neurons' activations as the compression variable when computing the MI. In their investigation of an uncalibrated fully connected neural architecture, they discovered that MI exhibited lower variance and higher values in the initial layers, with decreasing values and greater variance in the later layers. 
	
	\section{Information Bottleneck using Transfer Entropy} \label{IB-TE}
	Creating models that generalize well may necessitate large datasets and many training iterations. While some efforts have been made to determine the rate at which a network learns, such as assessing the optimal progression of the training process or detecting whether generalization decreases during training, these attempts are limited in number. We describe in this section our TE-based method to quantify information transfer between neural layers.
	
	Multiple metrics can be used to evaluate a neural network's performance, the most common being accuracy and number of parameters. Focusing only on accuracy and number of parameters can lead to a suboptimal training path or a less robust model, because we may omit  important features of the training process.  
	
	Our investigation is based on the assumption that it is feasible to quantify compression by calculating TE across neighboring layers. TE can be used to improve the latter but also as an adaptive parameter that can require less training epochs, as we shown in \cite{Moldovan2020, Moldovan2021}. 
	
\begin{figure}[htbp]
		\centerline{\includegraphics[width=85mm,scale=0.5]{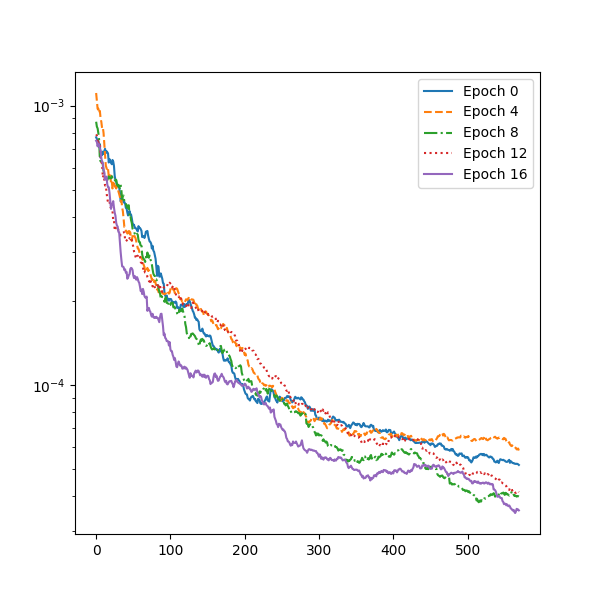}}
		\caption{TE evolution on selected equidistant epochs for FashionMNIST dataset on a logarithmic \emph{Y} axis. Each $Y$ value represents TE for a training sample. The $X$ axis shows the number of batches.}
		\label{fig:fashionepochs}
	\end{figure}

	We calculate TE using equation \ref{eq:TEcond}, where $i_t^{(k)}$ and $j_t^{(l)}$ refer to the series obtained from the binarized neuron activations, and $k$ and $l$ denote the neural network layers, in sequential order. Each training sample produces a single entry in the time series. By employing an optimized network architecture, we show that TE values exhibit higher values in the last layers and also higher variance. Within each epoch, TE gradually decreases, as depicted in Figures \ref{fig:ionosphereepochs} and \ref{fig:fashionepochs},  confirming that compression of irrelevant information is higher at the beginning of each epoch. Thereafter, the network progressively learns generic features and \emph{forgets} extraneous information. 
	
	The overall compressed information evolves incrementally during training and is unevenly distributed among the layers, with TE having highest values in the final layers. Previous research has typically constructed IP at the intersection of MI between $X$, $T$ respectively $Y$ and $T$, where $X$ and $Y$ denote the activations and outputs, we suggest that TE serves as a comparable representation for IP since it is obtained from two adjacent layers. By indirectly measuring compression using TE between adjacent layers, we obtain more concrete evidence of the fitting and compression phases at the layer level.
	
	\section{Experiments} \label{results}
	The experiments conducted in this study employ the neural architectures outlined in \cite{Moldovan2020, Moldovan2021}, namely shallow feedforward fully connected networks and CNNs. Prior to recording the TE values, we designed optimized architectures for each of the used datasets to obtain close to state-of-the-art accuracy values. To calculate TE during training, we bin the neuron activations on the observed layers, resulting time series for each neuron. For CNNs, the length of the time series is equivalent to the batch size, and it is limited for performance reasons; for the shallow networks TE is calculated for each training sample up to the epoch length.
	
	\subsection{Shallow architectures}
\begin{figure}[htbp]
		\centerline{\includegraphics[width=50mm,scale=0.5]{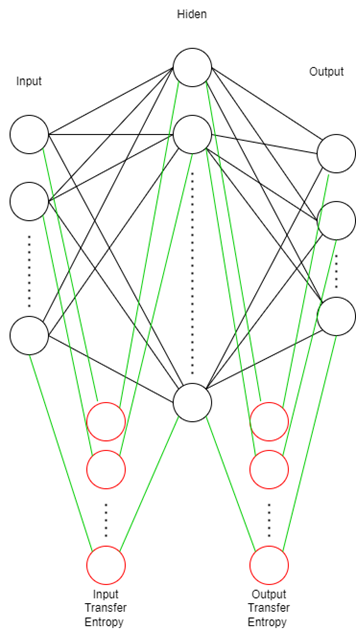}}
		\caption{Schematic of the three layer feedforward neural network. Each pair of layers contribute to obtaining a vector of TE values, illustrated with red circles, while green lines show the associated neuron pairs that produce the actual TE value.}
		\label{fig:ffnet}
	\end{figure}

	For the glass, ionosphere, seeds, divorce, and liver disorders datasets from UCI \cite{UCI2019}, we utilize an optimized single hidden layer network with the same structure and layer widths as described in Moldovan \emph{et al.} in \cite{Moldovan2020}. We compute TE for all adjacent layers and neurons, resulting two TE sets: "TE input" calculated between input and hidden layers and "TE output" between hidden and output layers. In order to allow neuron activation stabilization, we do not record activations for the initial 5\% training samples in each epoch. The binarization procedure uses a 95\% threshold percentile from each examined activation matrix. We emphasize the significance of utilizing a dynamic threshold since mean activation values change considerably during training. Our general layout for the feedforward network is illustrated in Figure \ref{fig:ffnet}.
	
	\begin{figure*}[htbp]
		\centerline{\includegraphics[width=155mm,scale=0.5]{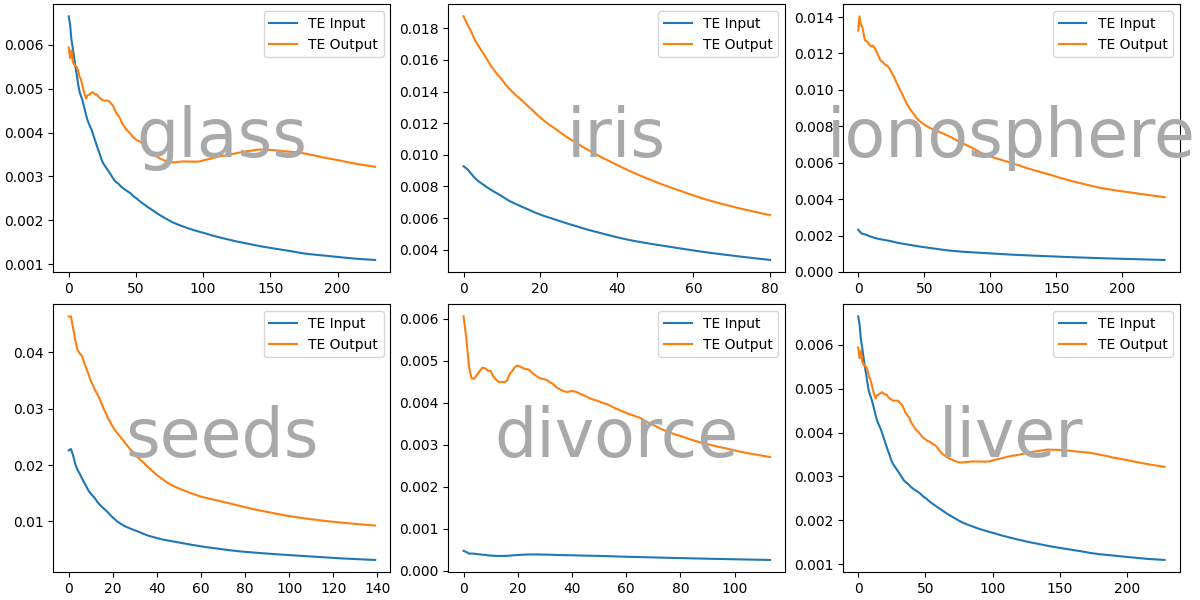}}
		\caption{Showing TE for a single hidden layer feedforward network. Each TE value plotted is the average TE for all training samples across all epochs. Epochs are 'stacked' along the \emph{x} axis. Observing \emph{glass} dataset results, we notice this is a hard to solve problem for single hidden layer networks, as it can be observed in Table \ref{datasets}; 'TE Output' evolution is slow and has small variance. }
		\label{fig:teinput}
	\end{figure*}
	
\begin{figure}[htbp]
		\centerline{\includegraphics[width=85mm,scale=0.5]{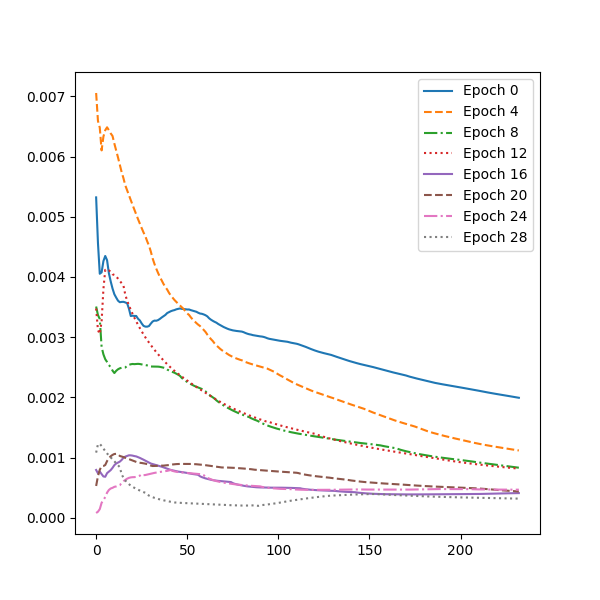}}
		\caption{Showing TE calculated between input and hidden layer, for each training sample, for every 4th epoch on the Ionosphere dataset. The number of epochs has been reduced to improve readability.}
		\label{fig:ionosphereepochs}
	\end{figure}

	Through analysis of the TE values for each training sample in every epoch, we observe a consistent trend where initial epochs exhibit higher TE values that gradually decrease over time. This pattern persists across all epochs, as illustrated in Figure \ref{fig:ionosphereepochs}, where the TE values show a continuous decrease from their initial values in subsequent epochs.
	
	In Figure \ref{fig:ionosphereepochs}, each line is associated with a specific epoch, where on average, higher TE lines correspond to lower epoch indexes. The $x$ axis represents the number of training samples available for a complete epoch. To gain an overall understanding of the TE trend, we average the TE array for each training sample across all epochs and reduce it to a single $y$ value as shown in Figure \ref{fig:teinput}. We believe that this provides a more precise representation of the fitting phase, as discussed by Shwartz-Ziv \emph{et al.} \cite{ShwartzZiv2017}. Towards the end of each epoch, the TE values show lower variation, which we associate with the \emph{compression} phase.
	
	\begin{figure*}[htbp]
		\centerline{\includegraphics[width=170mm]{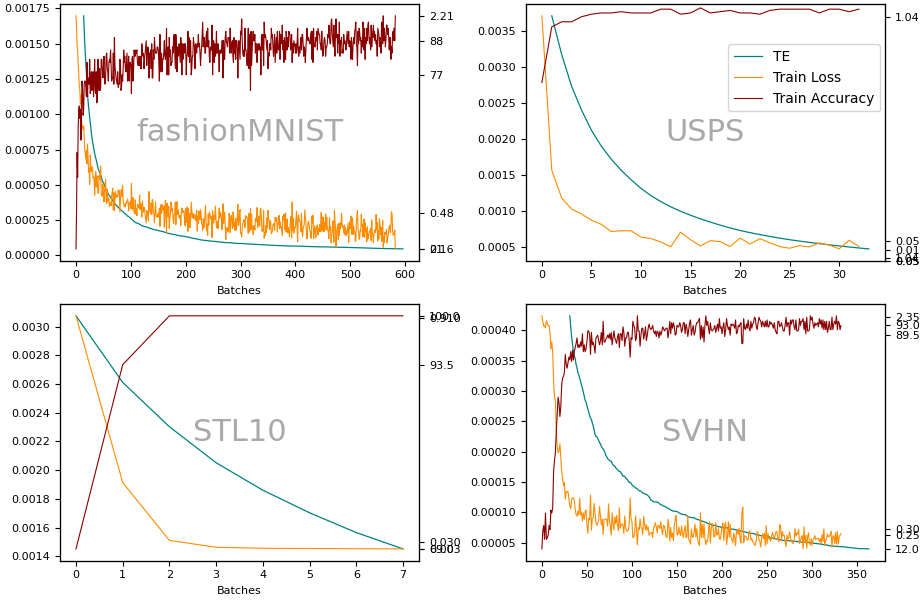}}
		\caption{Averaged by sample TEs for multiple networks for the last pair of layers (linear and softmax).}
		\label{fig:CNNmulti}
	\end{figure*}
	
\begin{figure}[htbp]
		\centerline{\includegraphics[width=85mm]{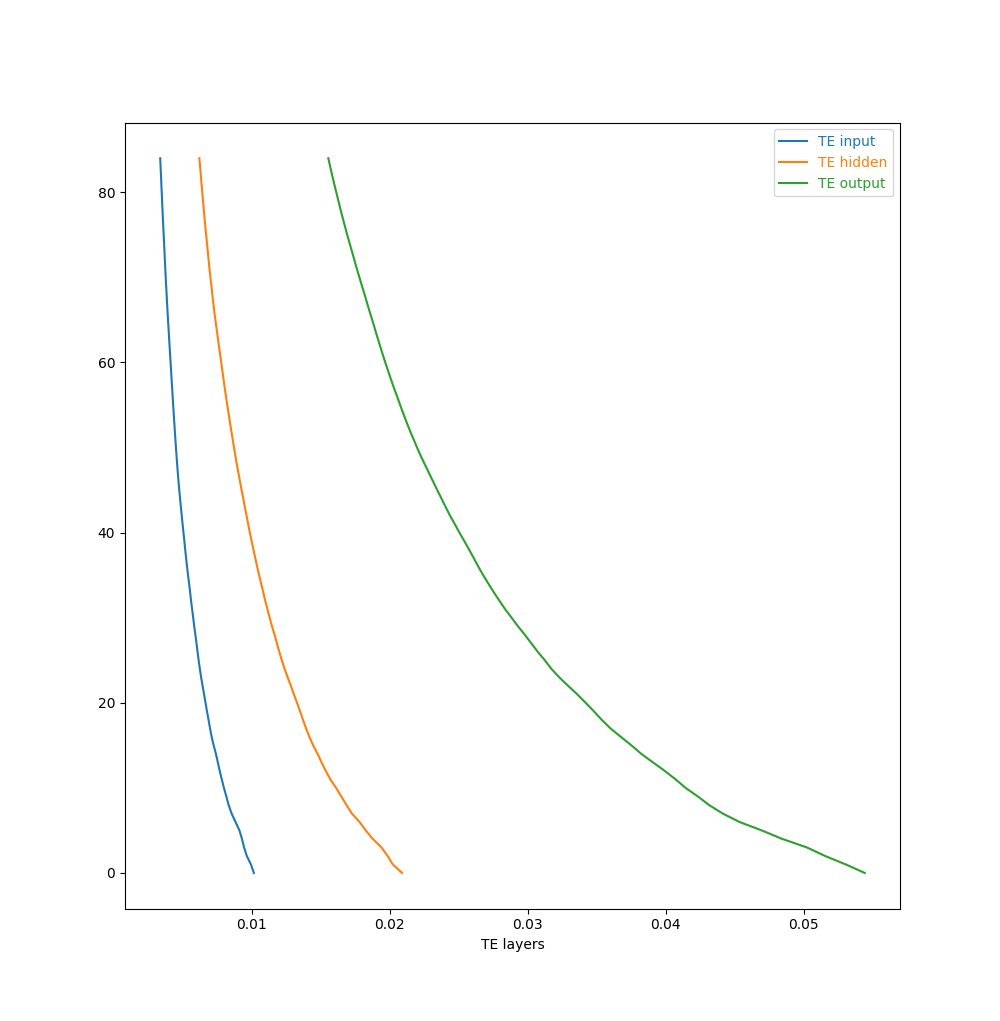}}
		\caption{Overtraining a fully connected network with 3 hidden layers on the Iris dataset while showing TE for, input to hidden layer, labeled as 'TE input', hidden to hidden and hidden to output marked with 'TE hidden' and 'TE output'. On the $x$ axis we plot the averaged normalized TE values for each training batch across all epochs.}
		\label{fig:iris3tes2}
	\end{figure}

	We conduct another experiment on a toy MLP fully connected architecture with 4, 8, 16, 8, respectively 3 neurons per layer. The first and last layers are the input and output layers. Due to computational constraints, we use the Iris dataset from UCI \cite{UCI2019}. To calculate TE, we employ a window of 20 training items. In each epoch, we skip the first 10 training items to allow for neuron activations to stabilize before using them in TE calculations. We observe that the averaged inter-layer TE lines match the IP from \cite{Saxe2019}, with the same sequencing: last layers produce larger TE values than the ones from the initial layers (Figure \ref{fig:iris3tes2}). This confirms that the last layers contain generic features, and receive most of the variations during training. Interestingly, even after overtraining the network up to 200 epochs, we do not observe a significant variation of the averaged TE values, despite achieving 97\% accuracy in just 10 epochs.
	
	
	\subsection{CNN architecture}
	In the second group of experiments, we utilize an optimized CNN,  based on the Alexnet architecture \cite{Alex2012}, for an image classification task on the following datasets: FashionMNIST \cite{fashionMNIST2017}, STL-10 \cite{STL102011}, SVHN \cite{SVHN2011}, and USPS \cite{uspsdataset}. We focus only on the last pair of fully connected layers (including the softmax layer), following the findings from \cite{Saxe2019}, that IB is present only in the fully connected layers. Another reason for the layer selection is the computational cost. However, the behavior of TE in these experiments is similar to that observed in shallow networks. For larger datasets, we also observe steeper but smoother trajectories in the averaged TE slopes, as shown in Figure \ref{fig:CNNmulti}. 

	In all experiments, we notice a direct correlation between the primary metrics of the network, such as accuracy and loss, and the changes in TE. Specifically, we observe that during the fitting phase there is a quick reduction of the TE values - that is similar to loss evolution and inverse to the accuracy, followed by an extended compression phase with consistently low TE variance.
	
	In our CNN experiments, TE follows the observations from \cite{ShwartzZiv2017,Tishby2015,Saxe2019}. The highest compression is found in the last fully connected layers. Since we evaluate TE after each training sample on shallow networks, we observe an immediate feedback in the TE value. The variability in TE's behavior is explainable by the used CNN architecture and the size of the datasets. Since Saxe \emph{et al.} \cite{Saxe2019} mentioned that SGD training does not yield better compression, we did not used it for shallow nets, but only for CNNs, due to performance reasons.
	
	\subsection{Discussion}
	TE decreases during and after each epoch because the compression rate 'slows down' as the patterns emerge and stabilize in each layer. Variances of TE during the fitting phase are mostly related to shuffling training samples for each epoch, as it is usually done in iterative training. At the same time, we empirically observe that TE is a sensitive metric to a network's architecture efficiency, determining various slopes and amplitudes during training. While choosing network parameters for experiments, we noticed that optimal architectures tend to produce smoother TE lines with a specific evolution pattern. Due to this,  we hypothesize that \emph{good} inter-layer compression is achieved only in efficient neural network structures. Studying TE at epoch level, we can visually depict on which epoch the network has reached a stable accuracy. In a similar fashion as Saxe's IP illustrates each layer's compression proficiency, TE dynamic visualization (at layer, epoch or training batch level) can depict possible hurdles during training. Figure \ref{fig:CNNmulti} shows that there is a strong inverse relation between train accuracy and TE, and also a close evolution of TE and calculated loss during training. On larger datasets, TE and training loss are closely following the same slope and relative magnitudes. Even if \cite{ShwartzZiv2017} correctly pointed out that measuring individual weights and neurons is meaningless, we see that averaging all TE values for a layer creates coherent TE lines during training.
	
	\section{Conclusion and Future Work} \label{conclusion}
	
	We can conclude that there is a strong connection between accuracy and loss on one side, and TE evolution on the other. The observations from our CNN experiments apply to smaller datasets and shallow architectures, but the latter are more susceptible to deficient network architectures and biased datasets. 
	
	Elad \emph{et al.} \cite{Elad2019} demonstrated that training a deep neural model layer-by-layer with an IB loss function based on MI results in comparable accuracy to end-to-end training using cross-entropy loss. However, some researchers have suggested using alternative cost functions instead of the IB loss, as it may not always behave optimally \cite{Amjad2020, Kolchinsky2019}. In our future work, we aim to explore the potential use of a TE-based loss function to enhance the training process of deep neural networks.
	
	\begin{table}[H]
		\caption{Performance on datasets}
		\centering
		\begin{tabular}{ccc}
			\toprule
			\textbf{Dataset}	& \textbf{Top 1 Accuracy}  &   \textbf{Epochs}\\
			\midrule
			fashionMNIST       & 88.96    &   22\\
			USPS  & 99.62\% &   10\\
			SVHN       & 92.30\%    &   10\\
			STL10       & 100.00\%  &   25\\
			glass     & 35.38\%    &   80\\
			iris  &   95.25\%   &   199\\
			ionosphere    &   95.28\%  &   30\\
			seeds &   84.12\%  &   100\\
			divorce   &   98.03\%   &   28\\
			liver &   64.42\%  &   57\\
			\bottomrule
		\end{tabular}
		\label{datasets}
	\end{table}

	\section*{Acknowledgment}
	All authors contributed equally to this article.
	This work was supported by Siemens SRL.

	\vspace{12pt}
	
\end{document}